\documentclass[sigconf]{acmart}
\AtBeginDocument{%
  \providecommand\BibTeX{{%
    \normalfont B\kern-0.5em{\scshape i\kern-0.25em b}\kern-0.8em\TeX}}}

\copyrightyear{2021} 
\acmYear{2021} 
\setcopyright{acmcopyright}
\acmConference[EAAMO '21]{Equity and Access in Algorithms, Mechanisms, and Optimization}{October 5--9, 2021}{--, NY, USA}
\acmBooktitle{Equity and Access in Algorithms, Mechanisms, and Optimization (EAAMO '21), October 5--9, 2021, --, NY, USA}
\acmPrice{15.00}
\acmDOI{10.1145/3465416.3483286}
\acmISBN{978-1-4503-8553-4/21/10}

\setcopyright{none}
\renewcommand\footnotetextcopyrightpermission[1]{}

\usepackage{xspace}
\newcommand*{\eg}{e.g.,\xspace}
\newcommand*{\ie}{i.e.,\xspace}

\usepackage{graphicx}
\usepackage{multirow}
\usepackage{subcaption}

\usepackage{stfloats}

\begin{document}

\title{Disaggregated Interventions to Reduce Inequality}

\author{Lucius E.J. Bynum}
\affiliation{%
  \institution{New York University}
  \streetaddress{60 5th Ave}
  \city{New York}
  \state{NY}
  \country{USA}
  \postcode{10011}
}
\email{lucius@nyu.edu}

\author{Joshua R. Loftus}
\orcid{0000-0002-2905-1632}
\affiliation{%
  \institution{London School of Economics}
  \streetaddress{Houghton Street}
  \city{London}
  \country{United Kingdom}}
\email{J.R.Loftus@lse.ac.uk}

\author{Julia Stoyanovich}
\affiliation{%
  \institution{New York University}
  \streetaddress{60 5th Ave}
  \city{New York}
  \state{NY}
  \country{USA}
  \postcode{10011}
}
\email{stoyanovich@nyu.edu}


\begin{abstract}
    A significant body of research in the data sciences considers unfair discrimination against social categories such as race or gender that could occur or be amplified as a result of algorithmic decisions. Simultaneously, real-world disparities continue to exist, even before algorithmic decisions are made. In this work, we draw on insights from the social sciences brought into the realm of causal modeling and constrained optimization, and develop a novel algorithmic framework for tackling pre-existing real-world disparities. The purpose of our framework, which we call the ``impact remediation framework,'' is to measure real-world disparities and discover the optimal intervention policies that could help improve equity or access to opportunity for those who are underserved with respect to an outcome of interest. 
    
    We develop a disaggregated approach to tackling pre-existing disparities that relaxes the typical set of assumptions required for the use of social categories in structural causal models. Our approach flexibly incorporates counterfactuals and is compatible with various ontological assumptions about the nature of social categories. We demonstrate impact remediation with a hypothetical case study and compare our disaggregated approach to an existing state-of-the-art approach, comparing its structure and resulting policy recommendations. In contrast to most work on optimal policy learning, we explore disparity reduction itself as an objective, explicitly focusing the power of algorithms on reducing inequality.
\end{abstract}



\maketitle

\pagestyle{plain}  

\section{Introduction}
Researchers have worked over the past several years to document and study how automated decision systems can systematically create and amplify unfair outcomes for individuals and groups. Recent work by \citet{impacts_kusner2019} has framed two complementary aspects of reducing unfair discrimination in algorithms: (1) the \emph{discriminatory prediction problem}, where we aim to define and reduce discrimination in model predictions; and (2) the \emph{discriminatory impact problem}, where we aim to define and reduce discrimination arising from the real-world impact of decisions. In both of these contexts, we are concerned with defining and mitigating unfair discrimination that could occur or be amplified as a result of an algorithm-assisted decision. Here, we introduce a third domain---one in which we still care about the potential biases of our model, but rather than focusing primarily on unfairness emerging from our model, we focus our modeling attention on disparity that \emph{already exists} in the world and is directly measured. Building on the discriminatory impact problem, we call this problem \textbf{impact remediation}, because the objective of our modeling process is to remediate an already-observed disparity via an intervention. 

In more concrete terms, we define impact remediation as follows: 
\begin{enumerate}
    \item We observe that there is an existing disparity between different groups of people with respect to an outcome in the world. The outcome in question is desirable. We consider this group-level disparity to be unfair.
    \item We have the ability to perform an intervention that decreases disparity between groups by improving outcomes for the disadvantaged groups.
    \item Interventions occur over groups of individuals, that is, a single intervention affects multiple individuals --- a fixed set of individuals --- depending on location or access with respect to the intervention. However, the sets on which we can intervene may constitute a different partitioning of people than the groups across which we observe disparity.
\end{enumerate}

There are several potential examples of impact remediation depending on the outcome in question, \eg people's access to education, voting, jobs, credit, housing, or vaccines. Oftentimes, we observe disparity across different groups of a \emph{social category} such as race or gender. For a concrete example, let's imagine that we are a hiring committee, and we observe a gender imbalance in our job applicant pool. Assume we know that each group wants to apply at roughly the same rate, so the problem is \emph{access} not interest. We thus want to increase access to our applicant pool across gender groups via outreach. Say we have an \emph{intervention} we can perform: we can choose where and how we host booths at career fairs. Here our \emph{impact} of interest could be the application rates for each gender group at each of the possible career fairs. The disparity we want to \emph{remediate} is the observed imbalance in application rates. 

How should we define disparity? Which career fairs should we target to get the best results? What resources would we need to spend in order to sufficiently improve access? In broad terms, these are the questions that our framework for impact remediation tackles via causal modeling and optimization. While these questions are not new, the novelty of the impact remediation problem is in formalizing how we can address these questions (1) with relaxed modeling assumptions about the nature of social categories; (2) in a multi-level fashion, where the groups of people on which we can intervene (\eg career fair attendees) are not necessarily the groups across which we observe disparate outcomes (\eg applicants by gender); and (3) while building on recent work that supports interference between interventional units during constrained optimization with causal models.

The impact remediation problem is directly tied to causal modeling because causal models systematize the study of interventions we may wish to perform. In recent years, causal modeling has emerged as a potential modeling tool for several questions of interest in algorithmic fairness and transparency, from defining fairness criteria to explaining classifications \cite{kusner2017counterfair, Kilbertus2017AvoidingDT, Chiappa2019PathSpecificCF, Zhang2018FairnessID, Karimi2020AlgorithmicRU}. These questions often involve social categories such as race and gender. With no universally agreed-upon specification of how social categories are lived and made real through experiences and discrimination, defining coherent counterfactuals with respect to indicators of social categories can be a significant challenge  \cite{kasirzadeh_smart_misuse_counterfact_2021, whats_sex_got_hu_hausmann_2020}. Further, the use of a social category is never divorced from its history. These two facts have led to an ongoing debate in the literature on an appropriate way to use social categories in causal models, and even on whether they should be used at all. 

Our framework relaxes the typical set of required modeling assumptions to use social categories with causal models, showing possible compatibility between areas of the literature on causal inference and social categories that would otherwise seem to be at odds. The purpose of social categorization in the impact remediation problem is, by design, to undo the social circumstances that make such categories relevant when they should not be. 

\paragraph{Contributions.} We make the following contributions:
\begin{itemize}
    \item We formalize the \textbf{impact remediation} problem using structural causal models.
    \item We present a framework for \textbf{using social categories in structural causal models} that relaxes the typical set of required modeling assumptions, showing possible compatibility between different sides of an ongoing debate in the literature.
    \item We present a hypothetical \textbf{case study with real data} in which we qualitatively and quantitatively illustrate the flexibility of our framework and contrast our approach to the state-of-the-art discriminatory impact problem from \citet{impacts_kusner2019}.
\end{itemize}

\section{Background \& Related Work}
\label{sec:background}

The impact remediation problem is aimed at generalizing a broad range of real-world problems. Consequently, the corresponding research context is also broad, including areas such as measurement, sociology, economic policy, causal inference, and optimization. We first provide some technical background on causal inference before discussing a few related works.

\paragraph{Causal Models and Counterfactuals.} Following \citeauthor{Pearl2000CausalityMR}, we use the structural causal model (SCM) framework. Using notation from \cite{Zhang2018FairnessID}, the framework defines an SCM as a 4-tuple ($U$, $V$, $F$, $P(U)$), where: $U$ is a set of exogenous background variables; $V$ is a set of endogenous observed variables, each of which is a descendant of at least one variable in $U$; $F$ is a set of functions (structural equations) defining, for each $V_i \in V$, the process by which it is assigned a value given the current values of $U$ and $V$; and $P(U)$ is a distribution over the exogenous variables $U$. Each SCM has a corresponding graphical model with $U, V$ represented as nodes and $F$ represented as directed edges between the associated input/output nodes, from causal parent to child in a directed acyclic graph (DAG). In short, the SCM fully specifies the data generating process, with each $V_i$ as a function of its parents in $V$ and any incoming exogenous variables. 

With notation from \cite{kusner2017counterfair, pearl2016primer}, we define an \emph{intervention} on variable $V_i$ as the substitution of equation $V_i = f_i(V, U)$ with a particular value $V_i = v$, simulating the forced setting of $V_i$ to $v$ by an external agent. In a fully specified SCM, we can use the distribution $P(U)$ over the exogenous variables $U$ to derive the distribution of any subset $Z \subseteq V$ following intervention on $V \setminus Z$ \cite{kusner2017counterfair}. For any $X \in V$, if we observe $X=x$ in our data, we can describe \emph{counterfactual} quantity $x'$ for $X$ as the value $X$ would have taken had $Z$ been $z$ in situation $U=u$, given that we observed the `factual' value $X=x$. We leave out further detail on how counterfactuals are estimated and refer the reader to \cite{pearl2016primer, Pearl2000CausalityMR} for an overview.

\paragraph{Interventions with Interference.} Following \cite{impacts_kusner2019, Sobel2006WhatDR,Ogburn2014CausalDF,aronow2017estimating}, we consider \emph{interference models}, which generalize SCMs to allow interventions on one unit $i$ to affect another unit $j$. Building directly on \citet{impacts_kusner2019}, we also focus only on intention-to-treat intervention effects, where, given a set of interventions $\{ Z^{(1)}, \ldots, Z^{(m)} \}$ and measured outcomes $\{ Y^{(1)}, \ldots, Y^{(m)} \}$ across $m$ units, interference models allow each outcome $Y^{(i)}$ to be a function of the entire set of interventions $\boldsymbol{z} \equiv [z^{(1)}, \ldots, z^{(m)}]^T$ rather than a function of $z^{(i)}$ only. Note that non-interference is still included as a special case. Recent work such as \cite{bipartite_interference_2021, Doudchenko2020CausalIW} has also studied interference in settings like ours, with interventions across sets of individuals and interconnected units.

\paragraph{Discriminatory Impacts and Fair Optimal Policies.} Impact remediation has a similar setup to the discriminatory impact problem introduced by \citet{impacts_kusner2019} and can be seen as an extension of their framework.  We use the same definition of an \emph{impact}: a real-world event caused in part by controllable algorithmic decisions as well as uncontrollable real-world factors that may themselves be biased. The goal in their framework is to maximize overall beneficial impact while accounting for any benefit that may be \emph{caused by} protected attributes $A$. We diverge from the setting of \citeauthor{impacts_kusner2019} in overall purpose as well as setup with respect to protected attributes. Where their problem is concerned with maximizing overall beneficial impact while estimating causal privilege, ours is instead aimed at minimizing measured disparity with optional in-group constraints that need not (but still can) estimate causal privilege. 

What may at first seem to be a subtle shift has large methodological consequences beyond the objective function of our corresponding optimization problem. One major implication of this shift in focus is that our framework here does not require using social categories as causal variables, so the definitions of social categories are separated from the evaluation of counterfactual statements. We can handle protected attributes flexibly depending on whether or not we believe they can be modeled as causes and have coherent counterfactuals. If we are interested in a particular protected attribute but believe it cannot be modeled as a cause, we do not include that attribute in $A$ and instead treat it as a social category across which we partition observed individuals (see Section~\ref{sec:social_categories} for details). Despite this separation, unfairness with respect to social categories is still a primary concern we are able to address --- such disparities are directly measured (\ie factual). To highlight the different methodological focus of our framework compared to that of \citet{impacts_kusner2019}, we demonstrate our approach in the same application domain as their example: allocating funding to NYC public schools. (See Section~\ref{sec:case_study} for details on this example.) Similar related works such as \cite{Nabi2019LearningOF, liu_delayed_2018, Green2019DisparateIA, Kannan2019DownstreamEO, Madras2018PredictRI} consider learning optimal fair policies and fairness in decision problems given real-world impacts, with our work differing in its focus on disparity rather than fairness, and in its alternate approach to modeling social categories.

\paragraph{Inequality and Merit} Our approach here also draws on the framing of algorithmic decision-making presented by \citet{fairness_equality_power_abebe_kasy_2021}, who raise the issue that algorithms concerned with fairness among those considered of equal `merit' by a decision-maker's objective can reinforce rather than question status quo inequalities. They propose alternative perspectives concerned with (1) the causal impact of decision-making algorithms on inequality and (2) the distribution of power and social welfare that can rationalize decisions or determine who gets to make them. 
Questions of inequality, social welfare, and optimal decision-making in light of these quantities are increasingly discussed in computing and algorithmic fairness \cite{abebe_2020_subsidy_shocks,Hu2020FairCA, Mullainathan2018AlgorithmicFA, Heidari2018FairnessBA, Heidari2019AMF} and relate to a vast literature in economics.
Drawing primarily on the perspectives in \cite{fairness_equality_power_abebe_kasy_2021}, which explore the tensions between a decision-maker's objectives, measures of fairness, and measures of inequality or social welfare, our focus here is complementary: we consider the case where the decision-maker's objectives and the measures of inequality or social welfare are one and the same.

\section{Representing Social Categories}\label{sec:social_categories}

When looking at disparities in the real world, it is often the case that we need to consider social categories. Representing social categories mathematically, however, can be difficult given how complex and multi-faceted they can be. This is especially true with causal modeling, in which we make additional assumptions about variables and the relationships between them in order to model cause and effect.

\begin{figure}
\begin{subfigure}[t]{0.225\textwidth}
\centering
\includegraphics[height=1.5cm]{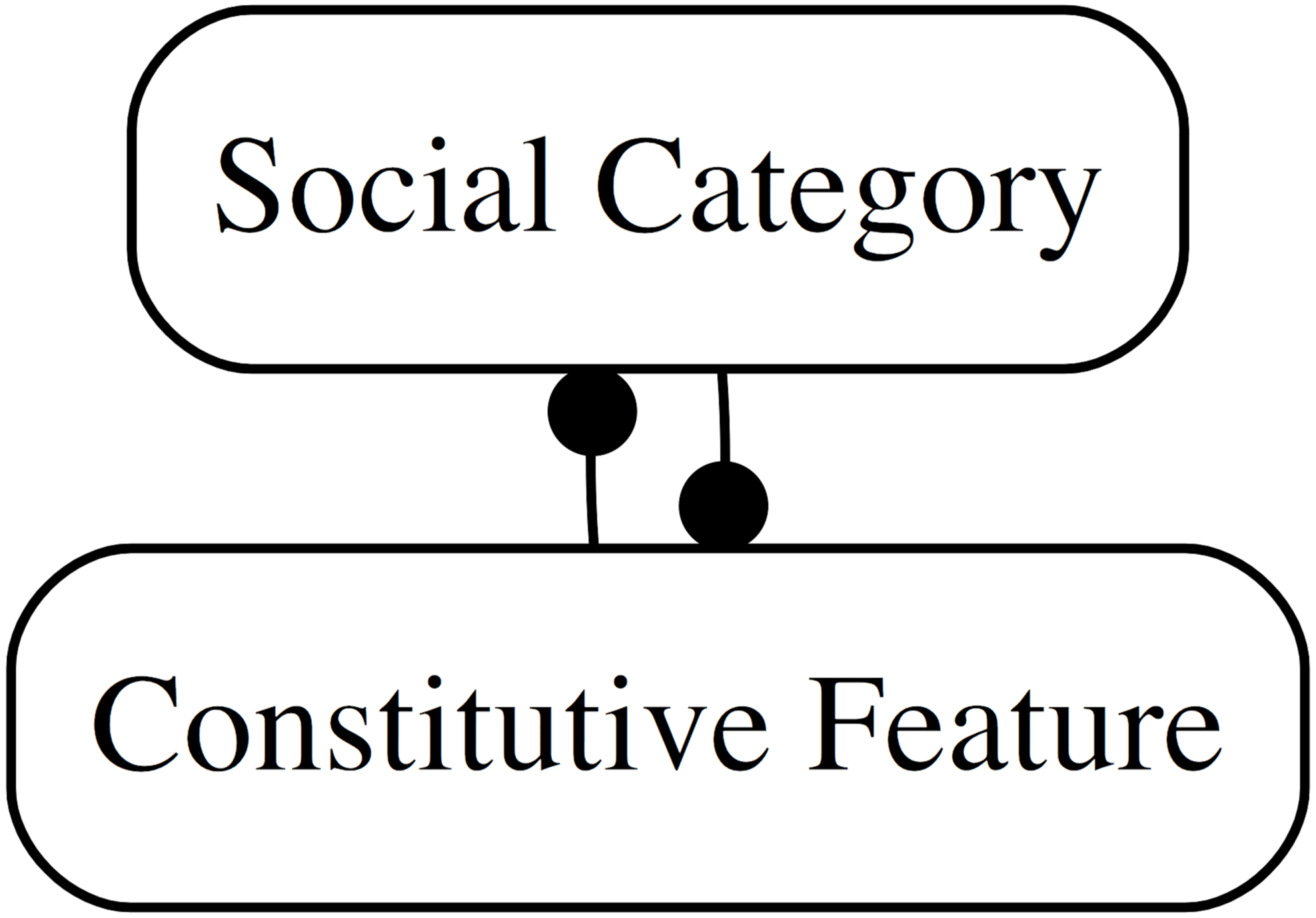}
\caption{ }
\label{fig:constitutive_category_graph}
\end{subfigure}
\begin{subfigure}[t]{0.225\textwidth}
\centering
\includegraphics[height=1.5cm]{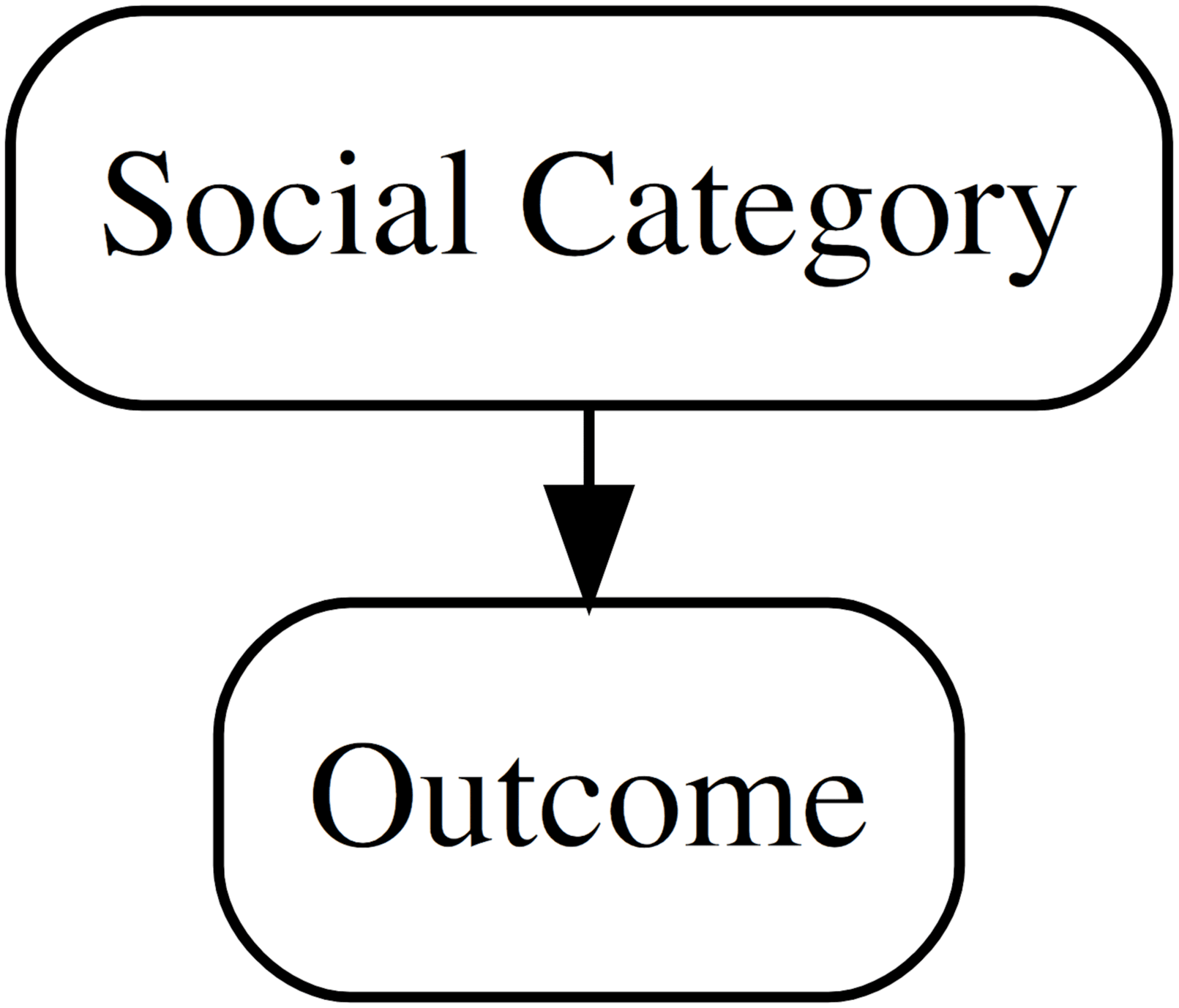}
\caption{ }
\label{fig:category_as_cause_graph}
\end{subfigure}
\caption{Two debated cases of the use of social categories in causal DAGs. We use circular arrowheads in DAG notation to denote constitutive rather than causal relationships. \textbf{(a)} In this DAG, we see that constitutive features are defining features of social categories. Rather than cause each other, they synchronously define each other, rendering the concept of modular cause-and-effect inapplicable if we choose to include two such variables in a causal model at the same time.  \textbf{(b)} This DAG exemplifies encoding a social category as a variable and modeling it as a cause for an arbitrary outcome. This model is debated depending on beliefs about whether an intervention on the social category in question can be modeled and/or is well-defined.}
\end{figure}

\paragraph{Social Categories as Variables and Causes.} To contextualize our approach, we find it useful to summarize a few specific methodological sources of debate related to the underlying assumptions of using social categories as treatments in causal modeling. Our discussion draws on a collection of previous works that wrestle with the use of social categories in machine learning and algorithmic fairness 
\cite{benthall_racial_2019, hanna_towards_2020, whats_sex_got_hu_hausmann_2020, kasirzadeh_smart_misuse_counterfact_2021, sen_race_2016, measurement_jacobs_wallach_2021}. 
We refer to a defining attribute of a social category as a \emph{constitutive feature}.
Constitutive relationships, to borrow language from \citet{whats_sex_got_hu_hausmann_2020}, are synchronous and definitional: instantaneously defining one piece of what it means to be in a social category; rather than causal and diachronic: unfolding over time via cause-and-effect. 
One point of debate present in discussions of social categories in causal models surrounds which features in a given problem we consider to be in constitutive relationships, if any, and how to incorporate such features in a causal DAG. 
If we choose to include two features that we consider instantaneously defining each other in a causal DAG, there is no way to properly define cause-and-effect with arrows. Figure~\ref{fig:constitutive_category_graph} illustrates this issue. We use circular arrowheads in DAG notation to denote constitutive rather than causal relationships. With two such nodes in a would-be DAG, we end up with an instantaneous constitutive cycle. Often the argument is about whether or not it is fundamentally inaccurate to remove a constitutive cycle via simplifying assumptions and draw a single arrow one way or the other. 
A feature pair that illustrates a potentially constitutive relationship is racial category paired with socioeconomic history, depending on our beliefs about the nature of the social category `race.'  These complex constructs might show up as simplified variables in our models, but the above issues of synchronicity between their definitions remain, often turning what may initially seem like a simple modeling problem into a serious sociological endeavor.

Apart from the case of including constitutive features and social categories in a DAG at the same time, encoding a social category as a variable and using it as a cause at all is also a source of debate. Figure~\ref{fig:category_as_cause_graph} depicts the use of a social category in a DAG as a cause for an arbitrary outcome. There are various pieces of the debate around this model, often discussed in the context of racial categories. \citet{sen_race_2016} outline three issues in the context of causal modeling with racial categories as potential treatments: (1) an inability to manipulate race; (2) technical issues of post-treatment bias, common support, and multicollinearity; and (3) fundamental instability in what the construct race means across treatments, observations, and time. As potential paths forward, \citeauthor{sen_race_2016} suggest (a) using exposure to a racial cue as treatment when looking for evidence of the effects of racial perception on behavior, or (b) measuring the effect of some manipulable constitutive feature of race in a within-racial-group study when looking for mechanisms behind disparate outcomes. More recent work by \citet{hanna_towards_2020} similarly discusses the multidimensionality of a concept like race and the limitations for algorithms on conceptualizing and modeling race if a multidimensional perspective is taken. Other recent work, such as \cite{whats_sex_got_hu_hausmann_2020, kasirzadeh_smart_misuse_counterfact_2021}, points out that unspecified assumptions about or oversimplification of social categories can lead to incoherent counterfactuals, arguing that perhaps the inherent complexity of social categories ultimately cannot be reduced in a way that makes them compatible with causal intervention.

\begin{figure}
\begin{subfigure}[b]{0.4\textwidth}
\centering
\includegraphics[height=1.5cm]{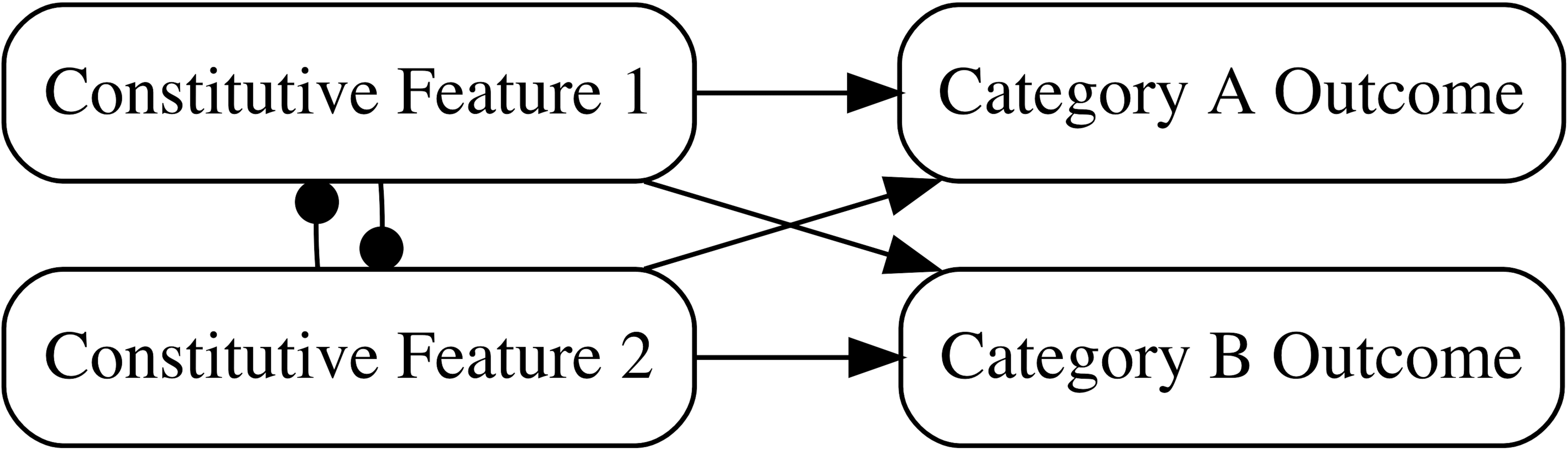}
\caption{ }
\label{fig:constitutive_pair_graph}
\end{subfigure}
\begin{subfigure}[b]{0.4\textwidth}
\centering
\includegraphics[height=1.5cm]{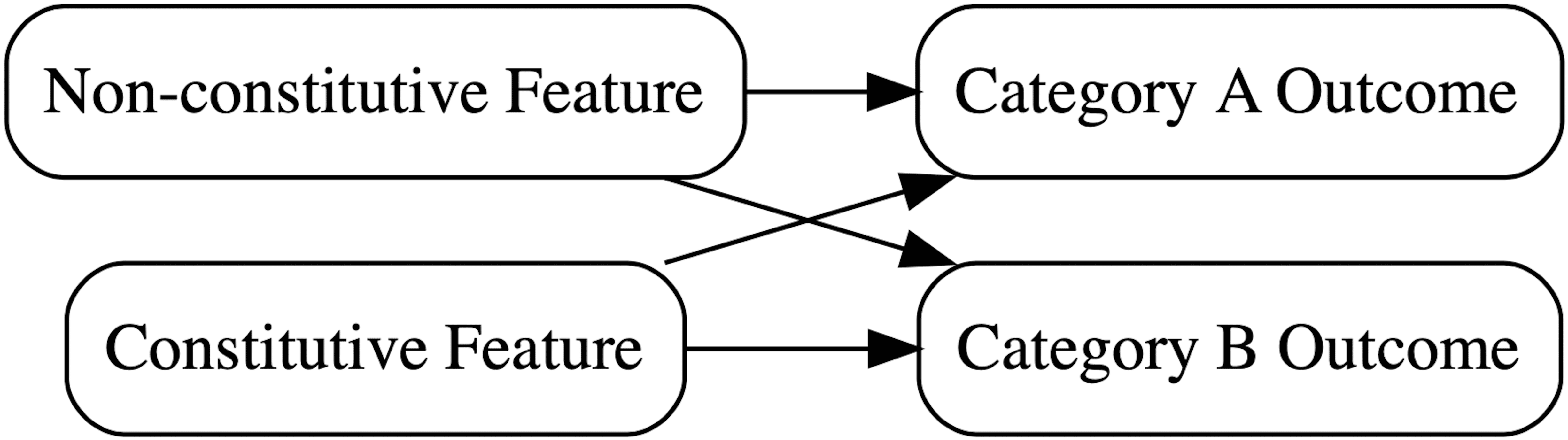}
\caption{ }
\label{fig:non_constitutive_graph}
\end{subfigure}
\caption{Two potential DAGs in the impact remediation problem. We use circular arrowheads in DAG notation to denote constitutive rather than causal relationships. \textbf{(a)} In this DAG we have two constitutive features that can synchronously affect or define each other due to their connection to a social category, again rendering the concept of modular cause-and-effect inapplicable if we choose to include both in our causal model at the same time. We would not be able to proceed with impact remediation given such a DAG. \textbf{(b)} This DAG demonstrates that avoiding social categories as variables and using at most one constitutive feature (on which we believe interventions can be modeled) allows us to avoid constitutive arrows as well as interventions on social categories as causes.}\label{fig:new-design-graph}
\end{figure}

\paragraph{Our Approach: Social Categories as Partitions.} We take dealing with the above issues for causal modeling with social categories as inspiration for an alternate modeling framework that avoids the sticking points --- including multiple constitutive features and encoding social categories as causal variables --- in the debate outlined above but still addresses a central and longstanding concern relevant to algorithmic fairness research: pre-existing bias \cite{Friedman1996BiasIC, stoyanovich_responsible_2020}. The underlying conceptual shift in specifying this framework is a shift in focus away from modeling the effects of demographic attributes and towards the effects of real-world interventions for each demographic group in a disaggregated manner.

The idea of disaggregation is also often discussed in connection with race (see \citet{hanna_towards_2020} for an overview from the perspective of algorithmic fairness). Disaggregation in the context of social categories can refer to disaggregation of the social categories themselves (\eg picking apart the different dimensions of a category). Defining a social category and specifying its constitutive features could be considered part of this process. Disaggregation can also refer to the consideration of different groups of a social category separately rather than interchangeably. Although these two meanings relate to each other, it is the second use of the term `disaggregated' that we primarily refer to when we say `disaggregated intervention design' in this work. Inspired by the within-racial-group design proposed by \citet{sen_race_2016}, the fact that our outcome is disaggregated across social categories allows us to consider intervention effects within each category and attempt to re-balance the outcome of interest across categories. 

Our modeling framework treats social categories mathematically as \emph{partitions} over individuals rather than individual attributes. By virtue of this choice, the required modeling assumption about social categories in our intervention design is the following: \emph{a social category consists of a group of people}. This group may or may not have any shared attributes. The categories themselves may be socially constructed, physically real, subject to change, or have no inherent meaning at all. The categories are only given meaning in our framework with respect to a measured outcome of interest across which we observe undesired disparity. Because of the widespread realities of structural discrimination, social categories such as race and gender may often be salient for revealing disparity across a particular outcome of interest. (See Section \ref{sec:measurement} for additional discussion about operationalizing social categories.) Even with social categories as partitions, we still have to avoid the issue of including multiple constitutive features in a causal model simultaneously, as shown in Figure~\ref{fig:constitutive_pair_graph}. Figure~\ref{fig:non_constitutive_graph} demonstrates how our disaggregated design can avoid constitutive relationships in addition to relaxing required modeling assumptions about social categories. Note that if interventions on a constitutive feature are also considered undefined, it can instead be treated as a social category. With such a design, we are able to consider our disparity of interest across social categories as we employ causal models to look for optimal interventions.

\section{Impact Remediation}
\label{sec:problem-statement}

The impact remediation problem includes measuring an existing disparity, defining potential interventions, and then finding the optimal set of units on which to perform an intervention. We first describe the measurement step before introducing notation for measured quantities.

\begin{figure*}
\centering
{
\setlength\fboxsep{5pt}
 \setlength\fboxrule{0.25pt}
 \fcolorbox{cyan!55!black}{cyan!5!white}{\begin{minipage}
    {\dimexpr\textwidth-2\fboxsep-2\fboxrule\relax}
    {\setlength\fboxsep{2pt}\fcolorbox{cyan!55!black}{white!5!white}{\textcolor{black}{Example: Outreach in a Job Applicant Pool} \hfill}}
    Consider $n$ university-age students across two universities that make up a fictitious job applicant pool. Imagine we observe gender imbalance in application rates across these $n$ students for gender groups $A$ and $B$. We wish to re-balance our applicant pool with respect to gender through outreach to potential job applicants. The intervention we are able to perform is doing targeted outreach via hosting booths at each university's career fair. Let $\{ X^{(1)}, X^{(2)} \}$ be the number of career counselors at each university, $\{ (Y^{(1)}_{A}, Y^{(1)}_{B}), (Y^{(2)}_{A}, Y^{(2)}_{B}) \}$ be the fraction of students of each gender at each university who apply for the job, and $Z^{(i)} = 1$ if we host a booth at the career fair of university $i$ and $0$ otherwise. Let $n^{(i)}_A$ be the number of students who identify with gender category $A$ at university $i$ and let $n^{(i)}_B$ be the same for category $B$. Let $n_A = n^{(1)}_A + n^{(2)}_A$ and $n_B = n^{(1)}_B + n^{(2)}_B$, counting the total number of students in each gender group. One possible measure of disparity in application rates between genders is the following:
    \begin{equation}\label{eq:disparity_measure}
        \delta = \left|\frac{1}{n_A} \sum_{i=1}^{2} n^{(i)}_A Y^{(i)}_{A} - \frac{1}{n_{B}} \sum_{i=1}^{2} n^{(i)}_B Y^{(i)}_{B} \right|,
    \end{equation}
    measuring the absolute difference in overall application rates between groups $A$ and $B$. For this example, consider 
    $$n^{(1)}_A = 100,  n^{(2)}_A = 75, n^{(1)}_B = 150, n^{(2)}_B = 100.$$
    
    \begin{minipage}{0.4\textwidth}
        \centering
        \fcolorbox{black}{cyan!5!white}{\includegraphics[width=0.9\textwidth]{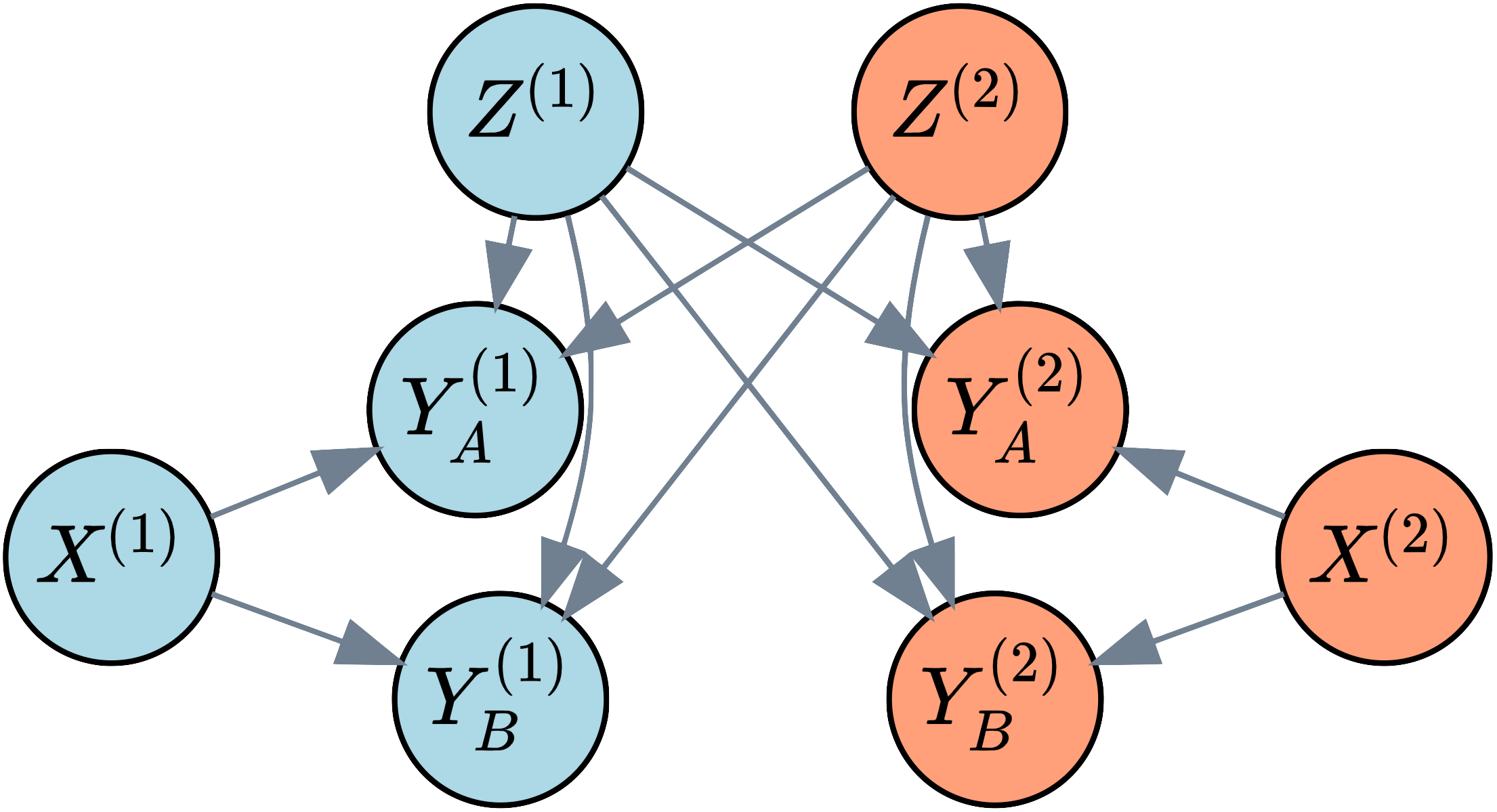}}
    \end{minipage}
    \hfill
    \begin{minipage}{0.58\textwidth}
        Let the graphical model on the left represent the causal relationship between $X, Y, Z$. Observe that we allow interference: imagine, for example, that these two universities partner to allow their students to attend either career fair. Given such a graphical model and its structural equations, we can estimate expected application rates after intervention $\mathbb{E}[Y^{(i)}_{k}(\boldsymbol{z})]$ for university $i$, social group $k \in \{A, B\}$, and set of interventions performed $\boldsymbol{z} \in \{0, 1\}^2$. We can replace $Y$ with $\mathbb{E}[Y(\boldsymbol{z})]$ in Equation \ref{eq:disparity_measure}, with $\delta(\boldsymbol{z})$ now denoting the measure of disparity \emph{after} intervention, a quantity we seek to minimize. Suppose that our measured application rates are the following:
        $$(Y^{(1)}_{A}, Y^{(1)}_{B}) = (0.10, 0.20), \hspace{0.5cm} (Y^{(2)}_{A}, Y^{(2)}_{B}) = (0.05, 0.10).$$
    \end{minipage}\\[0.05cm]
    
    Say we can host at most one career fair booth.  With no intervention, $\delta \approx 0.08$. From our causal model we obtain estimates of $\mathbb{E}[Y(\boldsymbol{z})]$ for both possible interventions:
    
    \begin{minipage}{0.5\textwidth}
    $$\mathbb{E}[Y^{(1)}_{A}([z^{(1)} = 1, z^{(2)} = 0])] = 0.20$$
        $$\mathbb{E}[Y^{(2)}_{A}([z^{(1)} = 1, z^{(2)} = 0])] = 0.10$$
        $$\mathbb{E}[Y^{(1)}_{B}([z^{(1)} = 1, z^{(2)} = 0])] = 0.30$$
        $$\mathbb{E}[Y^{(2)}_{B}([z^{(1)} = 1, z^{(2)} = 0])] = 0.15$$
    \end{minipage}
    \begin{minipage}{0.5\textwidth}
        $$\mathbb{E}[Y^{(1)}_{A}([z^{(1)} = 0, z^{(2)} = 1])] = 0.15$$
        $$\mathbb{E}[Y^{(2)}_{A}([z^{(1)} = 0, z^{(2)} = 1])] = 0.15$$
        $$\mathbb{E}[Y^{(1)}_{B}([z^{(1)} = 0, z^{(2)} = 1])] = 0.25$$
        $$\mathbb{E}[Y^{(2)}_{B}([z^{(1)} = 0, z^{(2)} = 1])] = 0.15$$
    \end{minipage}\\[0.05cm]
    
    from which we compute $\delta([z^{(1)} = 1, z^{(2)} = 0]) \approx 0.08$ and $\delta([z^{(1)} = 0, z^{(2)} = 1]) = 0.06$. Thus, we determine intervention $[z^{(1)} = 0, z^{(2)} = 1]$ to be the optimal way to do outreach for our applicant pool, and we host our career fair booth at university 2.
    
    \dotfill
    
    Observe that, in the above example, our focus is on finding the optimal choice of interventions given a constraint on our budget and a particular measure of disparity.  Alternatively, we could have used the same framework to compare multiple possible interventions, or to find the minimum budget that would allow us to achieve a specified disparity value, or even to compare multiple intervention decisions under different measures of disparity.  
    
 \end{minipage}}}
\end{figure*}

\subsection{Measurement and Intervention Formulation}
\label{sec:measurement}

As outlined in \citet{measurement_jacobs_wallach_2021}, measurement itself is a modeling process, a key part of which is the operationalization of constructs of interest as variables that can conceivably be measured. 

\paragraph{Operationalizing Categories of People.} In the measurement step, we may operationalize a social category. Notably, measurement can be used to obscure just as it can be used to reveal. As an example, if we put out a survey in which we operationalized race as a closed-ended self-classification \cite{hanna_towards_2020}, disparities we may or may not discover would depend on what categories we choose to include in the list. As another example, were we to operationalize gender as strictly binary, we may fail to measure outcomes for anyone who identifies in a non-binary way. For a given outcome, if we are thinking about intersectionality, we could choose to look at the Cartesian product of different categories. 

We emphasize that splitting outcomes across social category lines is context-dependent and not always necessary in the first place. Moreover, measurement of disparate outcomes here is not limited to traditional social categories. Depending on the application and the data we have, we could choose to partition people into social groups using other quantities, for example location, neighborhood, or even the outcome itself. A social partition based on the outcome itself could in some cases allow us to focus more exclusively on the mechanisms of an inequality, and is compatible with our framework. We highlight social categories such as race and gender in our presentation of this problem because of their relevance in many settings where discrimination has occurred historically. As discussed in Section~\ref{sec:social_categories}, the required modeling assumption about social categories in our framework---including those explicitly discussed---is only that they consist of groups of people, not necessarily with any shared attributes. 

\paragraph{Identifying Potential Interventions.} Alongside measurement of a disparate outcome, we require domain expertise to study the mechanisms by which disparity in our outcome of interest has arisen. Even without comprehensive knowledge of such mechanisms, which can be impossible to have in practical settings, we can still theorize potential interventions. For example, if a group of education researchers finds that one reason high school students have less access to college entrance exams is a shortage of Calculus teachers (even if that is not the only reason), providing funding for additional teachers is a potential intervention. That said, the better the intervention disrupts the actual mechanisms leading to disparity in the first place, the more effective it will be. 
While we believe users of this approach will generally be focused on beneficial interventions, we do not build this constraint into our modeling assumptions. Since it is possible for a well-intentioned intervention to be harmful to some individuals or groups, it may be important that a modeling constraint does not obscure that danger in cases where the data would otherwise highlight it.

In what follows, we present our proposed disaggregated intervention framework.  This framework will allow us to represent and quantify the effect of a proposed intervention, and to compare multiple possible interventions.

\subsection{Disaggregated Intervention Design}
\label{sec:design}

\begin{figure}
    \centering
    \fbox{\includegraphics[width=0.36\textwidth]{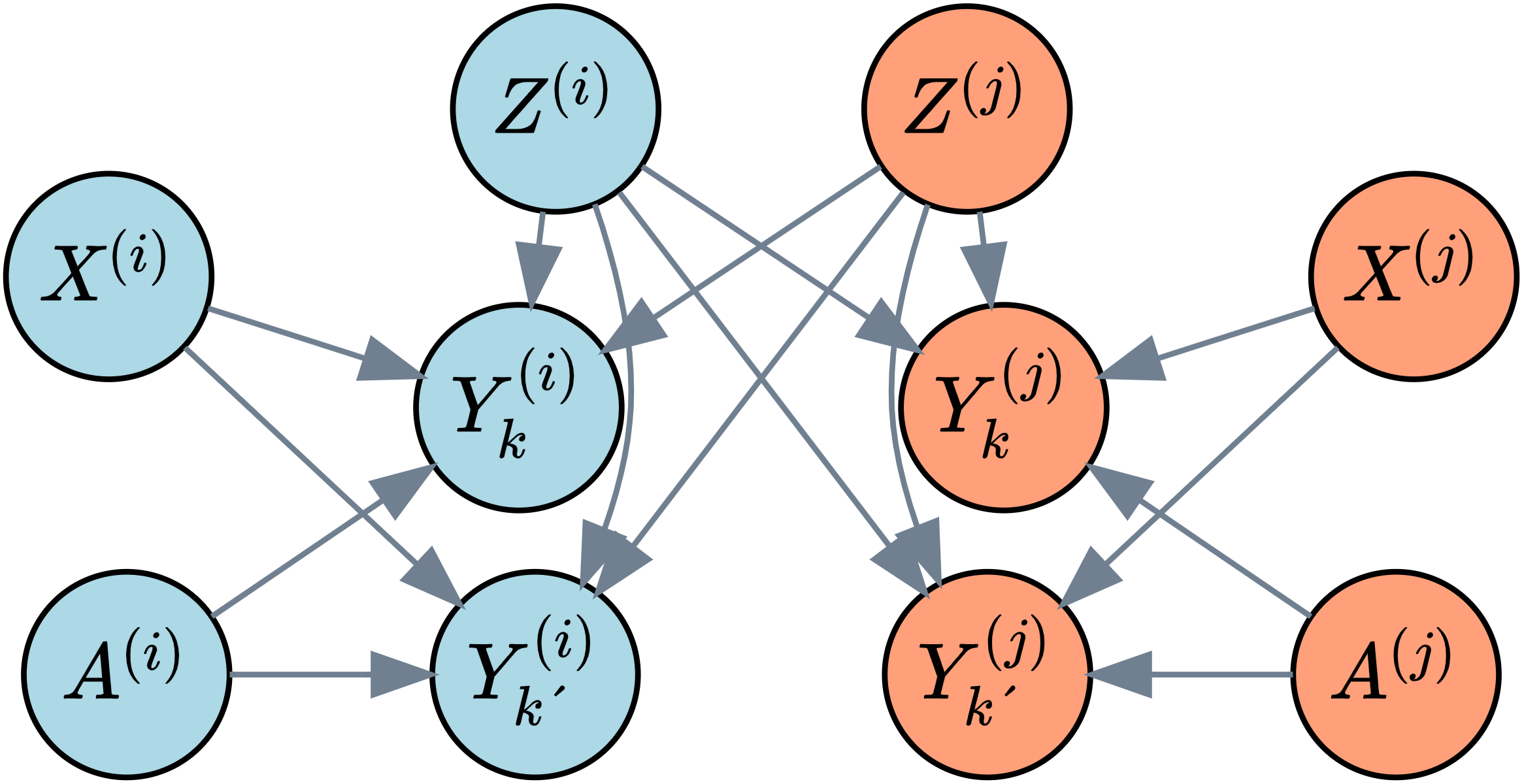}}
    \caption{An example causal DAG for the impact remediation problem with two intervention sets $i, j$ and two social categories $k, k'$. Variables $A, X, Z$ causally influence outcome $Y$, and intervention on one intervention set can potentially affect outcomes in the other, demonstrating interference. For example, $Z^{(i)}=z$ can potentially affect outcomes $Y^{(j)}_{k}, Y^{(j)}_{k'}$ in addition to $Y^{(i)}_{k}, Y^{(i)}_{k'}$.}
    \label{fig:archetypal_remediation_graph}
\end{figure}

We define our interventional optimization problem with the following notation. 
We have a set $I$ of $n$ individuals and a partition $\rho_Z$ of $I$ into $m$ \emph{intervention sets}: fixed sets of individuals affected directly by a single intervention. In other words, $\bigcup_{S_i \in \rho_Z} S_i = I$, $|\rho_Z| = m$, and $S_i \cap S_j = \emptyset$ for all $S_i, S_j \in \rho_Z$, where $i \neq j$. We also have another partition $\rho_C$ of set $I$ into $r$ sets representing each value of a social category or intersections (a Cartesian product) of several social categories. 
Let $n^{(i)}_k$ represent the number of individuals in both intervention set $S_i$ and social category $k$, where $n_k = \sum_{i=1}^{m} n^{(i)}_k$ captures the total number of individuals in social category $k$ and $n^{(i)} = \sum_{k=1}^{r} n^{(i)}_k$ captures the total number of individuals in intervention set $i$. We note in passing that these can be replaced with other weights $w^{(i)}_k$, $w_k$, $w^{(i)}$, respectively, in specific contexts where it makes sense to do so. In addition to the two partitions, we have an $(m \times d)$ matrix of real-world features $X$ for each intervention set, an $(m \times r)$ matrix of impact values $Y$ measured for each intervention set across the $r$ social category values, and an optional $(m \times \ell)$ matrix of protected attributes $A$ for each intervention set, where $A$ only includes attributes for which we decide that counterfactuals are well-defined. We have also identified a potential intervention $Z$ that can be performed on each of the $m$ intervention sets. We assume a causal graph describing how $A, X, Y, Z$ are  related. Figure \ref{fig:archetypal_remediation_graph} shows an example causal graph for the disaggregated design, similar to the discriminatory impact problem of \citet{impacts_kusner2019} but now with the variable $Y$ disaggregated across social categories in the causal graph. In each intervention set, all variables influence outcomes $Y$ for both of the social categories $k$ and $k'$. In addition, there is interference: intervention on one set can potentially affect outcomes in the other, for example with $Z^{(i)}$ influencing $Y^{(j)}_k, Y^{(j)}_{k'}$ as well as $Y^{(i)}_k, Y^{(i)}_{k'}$.

\paragraph{Model Fitting.} Our process involves several stages which roughly correspond to (1) modeling choices, (2) model fitting or estimation, and (3) optimization of the intervention. Modeling choices include structural assumptions built into the causal model as well as parametric or other functional forms for probability distributions and for relationships between observed and modeled-unobserved variables. In the estimation step, we choose an appropriate algorithm for fitting the assumed model to the observed data, and this algorithm estimates unknown parameters or functions which are then used as inputs into the final step. The estimation step may involve optimization, using any of a variety of statistical machine learning approaches, but in the current work when we speak of optimization we are referring to the intervention. Note that in this step we are not only learning a conditional expectation function to predict $Y$, but a function that also predicts $Y^{(i)}_k(\boldsymbol{z})$ under interventions $\boldsymbol{z}$. Finally, in the last step we use the fitted model to find an intervention by optimizing an objective function that reduces disparities.

During estimation and optimization, we must choose which covariates to include in the conditioning of various expectation functions. This is a subtle issue with no general solution, and some argue it requires causal knowledge to answer correctly \cite{Pearl2000CausalityMR}. We state simple guiding principles, but stress that these are not universal rules and that specific applications may justify other choices. 
Reasoning backwards from the goal of finding an optimal intervention, we need an accurate model because it influences the optimization problem. We could choose to condition on covariates when estimating the conditional expectation and intervention function $Y^{(i)}_k(\boldsymbol{z})$, for example, to negotiate a favorable trade-off between bias and variance so that the optimal intervention reduces actual disparities, and is not influenced by large errors due to a model that is underfit or overfit. On the other hand, we may not want our measure of disparity to be conditional on the same set of covariates as used in the estimation stage. This could be motivated by normative assumptions or considerations, for example whether that covariate is `resolving' of the disparity in the sense of \cite{Kilbertus2017AvoidingDT}. If necessary, we may be able to remove conditional covariates in the optimization step by marginalization.
If a covariate $P$ measures or correlates (potentially negatively) with privilege or opportunity, say---something which is considered a source of unfairness---then we should consider conditioning on $P$ for the purpose of reducing disparity. We may choose to do this independently of whether it would improve the model fit in the estimation stage. 

For brevity in this section, we notate expectations without conditioning, but we emphasize that if conditioning is done during the estimation step, we would also condition in the optimization step so as to optimize the same treatment function we estimated. Section~\ref{sec:case_study} shows an example with conditioning in both steps. In this paper, we demonstrate with particular choices at each of the stages for concreteness. In general, these choices should be made on an application-specific basis which could incorporate domain knowledge and constraints.

\paragraph{Variations on the Objective Function.}  It is often easier to observe unfair or undesirable circumstances than to define fairness at generality. The starting point for impact remediation is the observation and measurement of an imbalance in impact $Y$ that is considered unfair or undesirable. The situation-specific notion of fairness used within our framework is flexible, depending on what we consider a `fair' re-balancing of $Y$ would look like relative to the measured distribution. In the following sections, we consider $Y$ measured as a rate. Below are a few alternatives.

If we want the same rate across sub-populations indexed $k \in \{1, \ldots, r\}$, 
we might minimize the expression in Equation \ref{eq:within-objective} to enforce equal sub-population rates within each intervention set, 

\begin{equation}\label{eq:within-objective}
    \begin{aligned}
        \sum_{i=1}^{m} \sum_{k \neq k'} \left|\mathbb{E}[Y^{(i)}_{k}(\boldsymbol{z}) - Y^{(i)}_{k'}(\boldsymbol{z})] \right|,
    \end{aligned}
\end{equation}

\noindent or the expression in Equation \ref{eq:across-objective} to enforce the same sub-population rates across the entire population,

\begin{equation}\label{eq:across-objective}
    \begin{aligned}
        \sum_{k \neq k'} \left| \frac{1}{n_k} \sum_{i=1}^{m} n^{(i)}_k \mathbb{E}[Y^{(i)}_{k}(\boldsymbol{z})] -  \frac{1}{n_{k'}} \sum_{i=1}^{m} n^{(i)}_{k'} \mathbb{E}[Y^{(i)}_{k'}(\boldsymbol{z})] \right|.
    \end{aligned}
\end{equation}

The assumption in Equations \ref{eq:within-objective} and \ref{eq:across-objective} is that the outcome $Y$ is equally desired across groups, so the problem is lack of access.

If we instead believe that disparity between sub-population rates is okay once each rate is above a minimum threshold $\kappa$, we could analogously minimize the expression in Equation \ref{eq:min-within-objective} for enforcement within intervention sets,

\begin{equation}\label{eq:min-within-objective}
    \begin{aligned}
        \sum_{i=1}^{m} \sum_{k=1}^{r} \texttt{max} \left( \kappa - \mathbb{E}[Y^{(i)}_{k}(\boldsymbol{z})], 0 \right),
    \end{aligned}
\end{equation}

\noindent or Equation \ref{eq:min-across-objective} for enforcement across the population,

\begin{equation}\label{eq:min-across-objective}
    \begin{aligned}
        \sum_{k=1}^{r} \texttt{max} \left( \kappa - \left( \frac{1}{n_k}  \sum_{i=1}^{m} n^{(i)}_k \mathbb{E}[Y^{(i)}_{k}(\boldsymbol{z})] \right), 0 \right).
    \end{aligned}
\end{equation}

There are other variations as well, such as constraining the maximum gap between any two sub-populations, where we hold that a certain amount of disparity in rates is okay, or using expressions similar to those above but with different distance norms. In addition to any parameter thresholds we set, whether or not we can re-balance $Y$ will depend on what budget we have, the demographic distributions of the intervention sets, and the relative effectiveness of the intervention for each sub-population. Given these trade-offs, the impact remediation framework can be used to find optimal intervention sets with a fixed budget or to find the necessary budget given desired disparity thresholds: notice we could alternatively enforce Equations \ref{eq:min-within-objective} and \ref{eq:min-across-objective} as constraints while minimizing the intervention budget to determine the minimum budget required to raise each $\mathbb{E}[Y^{(i)}_{k}(\boldsymbol{z})]$ or $\frac{1}{n_k} \sum_{i=1}^{m} n^{(i)}_k \mathbb{E}[Y^{(i)}_{k}(\boldsymbol{z})]$ value above the desired threshold. The same setup can also be used to explore and identify feasible trade-offs for the given problem at hand.

\paragraph{Possible Constraints.} The constraints used during optimization will similarly be situation-specific. Possible constraints include the following.

\begin{itemize}
    \item A budget constraint, where we can afford at most $b$ interventions.
        $$\sum_{i=1}^{m} z^{(i)} \leq b$$
    \item A `do no harm' constraint, where we only allow a difference greater than or equal to $\eta$ (which could be zero) in expected outcome $Y$ before and after intervention. As with our possible objectives, this quantity could be measured within or across groups. For an example within each intervention set, we could have
        $$\mathbb{E}[Y^{(i)}_{k}(\boldsymbol{z})] - \mathbb{E}[Y^{(i)}_{k}] \geq \eta \ \ \forall (i, k) \in G$$
    for some subset $G$ of intervention sets and social groups. Alternatively, we could enforce the following constraint across the entire population:
        $$\frac{1}{n_k}  \sum_{i=1}^{m} n^{(i)}_k \left( \mathbb{E}[Y^{(i)}_{k}(\boldsymbol{z})] - \mathbb{E}[Y^{(i)}_{k}] \right) \geq \eta \ \ \forall k.$$
    Such a constraint can be important in the case that outcomes for one or more groups are decreased in order to decrease disparity between groups. See Section~\ref{sec:case_study} for an example.
    \item A fairness constraint involving protected attribute $A$. Such a constraint could be causal or associational. For a causal example, consider a $\tau$-controlled counterfactual privilege constraint, as defined in \citet{impacts_kusner2019}, where we constrain the counterfactual gain in $Y$ due to intervention on $A$:
        $$\mathbb{E}[Y^{(i)}_{k}(a^{(i)}, \boldsymbol{z})] - \mathbb{E}[Y^{(i)}_{k}(a', \boldsymbol{z})] < \tau, \ \ \forall a',i, k.$$
    \item As mentioned in our discussion of objective functions, we could enforce as a constraint that each social group rate be above a minimum threshold $\kappa$. For example, analogous to Equation \ref{eq:min-across-objective}, we could have
    $$\frac{1}{n_k}  \sum_{i=1}^{m} n^{(i)}_k \mathbb{E}[Y^{(i)}_{k}(\boldsymbol{z})] \geq \kappa.$$
\end{itemize}

As with optimization objectives, the possibilities for constraints are numerous beyond the set of examples we explore here. In short, the impact remediation framework's addition of disaggregation to constrained optimization with causal models is general enough to be compatible with a variety of problems. We now illustrate the use of our framework for a hypothetical use case.
 
 \section{Case Study: NYC Public Schools}\label{sec:case_study}
 
To showcase the practical usefulness of our framework and highlight the difference in methodological focus compared to that of \citet{impacts_kusner2019}, we demonstrate our approach in the same application domain as their example: allocating funding to NYC public schools. The original example problem imagines the US DOE intervening in the NYC public school system to offer funding for schools to hire Calculus teachers with the goal of increasing college attendance, as measured by the fraction of students taking college entrance exams. Our adaptation of this problem uses the same hypothetical intervention (allocating funding for schools to hire Calculus teachers) and the same outcome to facilitate comparison. We can imagine the outcome---the fraction of students taking college entrance exams---to be a proxy measure of access to higher education. This response is not the only measure we would want to use were we actually carrying out these interventions: whether or not a student takes a standardized test is not a comprehensive measure of their access to higher education, though, as in \citet{impacts_kusner2019}, it serves as a tangible example.

We will see that our disaggregated intervention design, given its different modeling strategy and objective, is not only different in design and required causal assumptions but can also result in a completely different set of interventions than the $\tau$-controlled counterfactual privilege approach of \citeauthor{impacts_kusner2019}. As mentioned above, a key step in impact remediation is the measurement of an existing disparity disaggregated across categories of interest. This need for disaggregation means that we need to collect a new, disaggregated dataset in order to study this problem. We collect data for $m=490$ public schools in NYC counties for the 2017-2018 school year from publicly available National Center for Education Statistics (NCES) and Civil Rights Data Collection (CRDC) data releases, and compare the impact remediation problem and discriminatory impact problem on this new dataset. \footnote{Our corresponding code repository can be found at \url{https://github.com/lbynum/reducing_inequality}.}

Note that in this hypothetical example, our focus is not on the causal models themselves, but instead on demonstrating the structure of impact remediation in a realistic setting. As such, we emphasize that the results and effectiveness of impact remediation are sensitive to our modeling choices as well as the accuracy and fit of each causal model, but we leave a rigorous sensitivity analysis of impact remediation to follow-on work.

\subsection{Original NYC Schools Example}\label{sec:schools}

To see directly how disaggregation differs from the approach of \citet{impacts_kusner2019}, we first introduce their original features of interest, causal model, and optimization framework. 

The original problem's variables of interest are as follows:

\begin{itemize}
    \item $R^{(i)}$ = distribution of CRDC racial/ethnic categories at school $i$ (set $R$ denotes all categories);
    \item $F^{(i)}$ = number of full-time counselors at school $i$ (fractional indicates part-time);
    \item $P^{(i)}$ = if school $i$ offers AP or IB courses (binary);
    \item $C^{(i)}$ = if school $i$ offers Calculus (binary); and
    \item $Y^{(i)}$ = fraction of students at school $i$ who take college entrance exams (SAT or ACT).
\end{itemize}

In the CRDC data release, the titles of the seven pre-defined racial categories in $R^{(i)}$ are the following, defined in detail in contemporaneous NCES report \citet{de2019status}: Black or African American (A), Hispanic or Latino (B), White (C), Asian (D), Native Hawaiian or Other Pacific Islander (E), American Indian or Alaska Native (F), Two or More Races (G). Emphasizing our modeling of these categories as one possible partition as well as our focus on disparity in outcomes rather than on the categories themselves, we denote them A through G in our analysis, as lettered above.

The assumed causal model is based on the following assumptions, with a stylized interference assumption:

\begin{itemize}
    \item All observed variables at school $i$ directly influence impact $Y^{(i)}$; and
    \item Students can take AP/IB courses or Calculus courses from any of the nearest $K=5$ schools to the school they attend, so $C^{(i)}$ and $P^{(i)}$ also influence $Y^{(j)}$ if school $j$ is one of the nearest $K$ schools to school $i$.
\end{itemize}

\begin{figure}
    \centering
    \fbox{\includegraphics[width=0.36\textwidth]{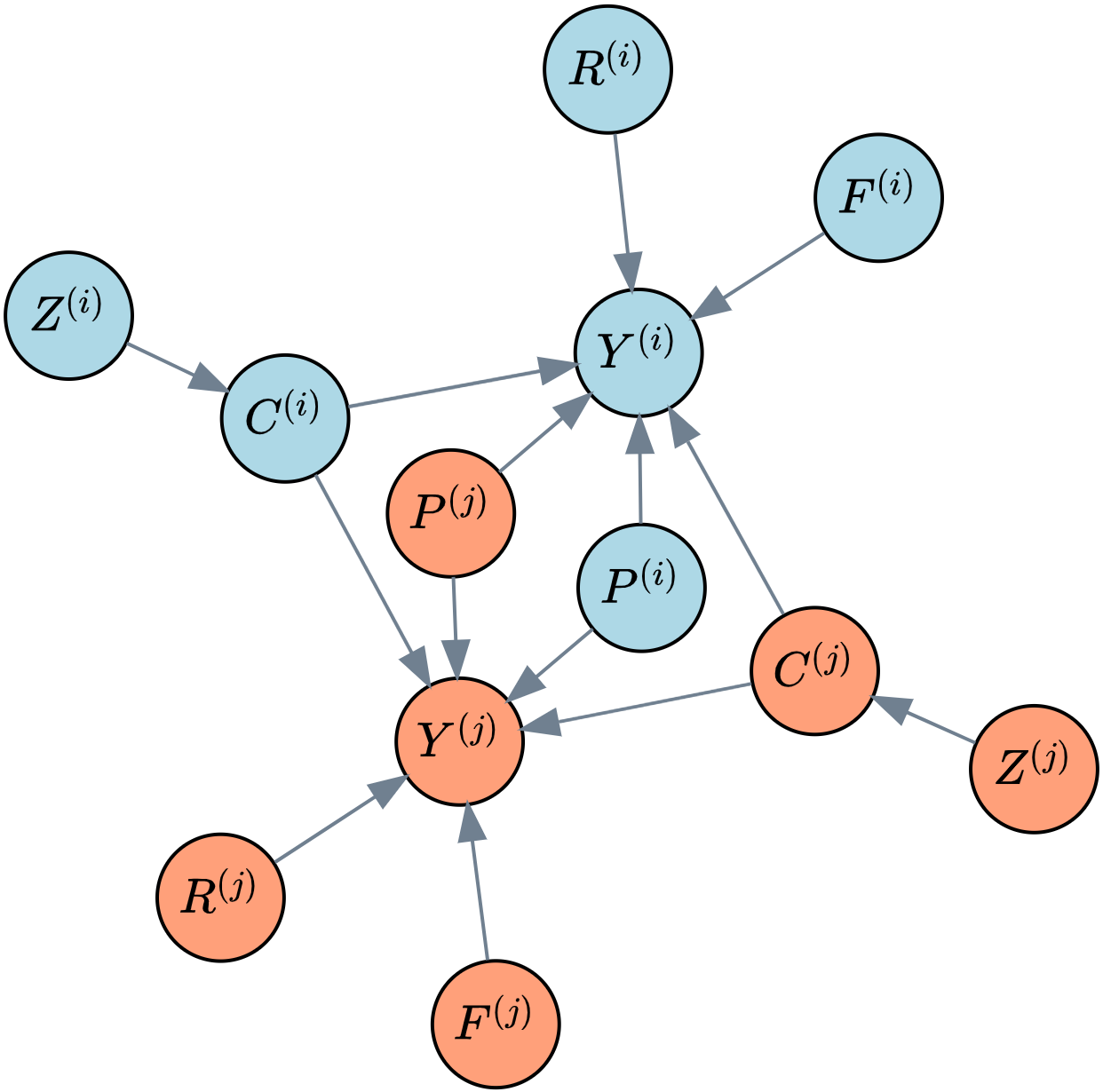}}
    \caption{Original causal graph structure for the NYC schools example from \cite{impacts_kusner2019}. This image depicts the subset of the graph corresponding to two neighboring schools $i$ and $j$, with outcome $Y^{(i)}$ causally influenced by $F^{(i)}, P^{(i)}, P^{(j)}, R^{(i)}$, and by interventions $Z^{(i)}, Z^{(j)}$ through $C^{(i)}, C^{(j)}$.}
    \label{fig:original_schools_graph}
\end{figure}

The corresponding causal graph structure for a pair of neighboring schools $i,j$ with interference is shown in Figure \ref{fig:original_schools_graph}. The full structural equations used for $Y$ are given by:

\begin{align*}
&\mathbb{E}[Y^{(i)}(\boldsymbol{r}, \boldsymbol{z})|R^{(i)} =\boldsymbol{r}^{(i)}, P^{(i)}=p^{(i)}, F^{(i)}=f^{(i)}]\\
&\hspace{0.5cm}= \boldsymbol{\alpha}^T\boldsymbol{r} \cdot \underset{\underset{\texttt{s.t.}, z^{(j)}=1}{j \in N(i)}}{\texttt{max}} s(i, j) C^{(j)}(\boldsymbol{z})\\
&\hspace{0.5cm}\hspace{0.1cm}+ \boldsymbol{\beta}^T\boldsymbol{r} \cdot \underset{j \in N(i)}{\texttt{max}} s(i, j) p^{(j)}\\
&\hspace{0.5cm}+ \boldsymbol{\gamma}^T\boldsymbol{r} f^{(i)}\\
&\hspace{0.5cm}+ \boldsymbol{\theta}^T\boldsymbol{r}
\end{align*}

where

\begin{itemize}
    \item $C^{(j)}(\boldsymbol{z}) = z^{(j)}$;
    \item $s(i, j)$ = inverse GIS distance between schools $i, j$;
    \item $N(i)$ = a set of the nearest $K=5$ schools to $i$ and also $i$ itself (a 6 element set);
    \item $\boldsymbol{r}^{(i)}$ = observed proportions of racial/ethnic categories at school $i$;
    \item $\boldsymbol{r}$ = any counterfactual setting of racial/ethnic category proportions at school $i$;
    \item $\boldsymbol{\alpha}, \boldsymbol{\beta}, \boldsymbol{\gamma}, \boldsymbol{\theta}$ are the structural equation parameters fit via least squares; and
    \item $\{\boldsymbol{r}^{(i)}, c^{(i)}, p^{(i)}, f^{(i)}, y^{(i)} \}_{i=1}^m$ are our observed data. \footnote{In slight contrast to the full structural equations in \citet{impacts_kusner2019}, in our second term involving $p^{(j)}$ we maximize over all $j \in N(i)$ regardless of whether $z^{(j)}=1$ because the intervention is specific to Calculus courses rather than all AP/IB courses.}
\end{itemize}

The purpose of the optimization in \citet{impacts_kusner2019} is to maximize the overall beneficial impact of an intervention, while constraining the estimated counterfactual benefit at school $i$ due to racial discrimination. \citeauthor{impacts_kusner2019} measure discrimination due to race via the notion of \emph{counterfactual privilege}. In this example, counterfactual privilege $c_{ir'}$ is defined as:

\begin{align*}
c_{ir'} &= \mathbb{E}[Y^{(i)}(\boldsymbol{r}^{(i)}, \boldsymbol{z})|R^{(i)} =\boldsymbol{r}^{(i)}, P^{(i)}=p^{(i)}, F^{(i)}=f^{(i)}]\\ &-\mathbb{E}[Y^{(i)}(\boldsymbol{r}', \boldsymbol{z})|R^{(i)} =\boldsymbol{r}^{(i)}, P^{(i)}=p^{(i)}, F^{(i)}=f^{(i)}]
\end{align*}

\noindent measuring the difference between the \emph{factual} benefit received by school $i$ after interventions $\boldsymbol{z}$ and the \emph{counterfactual} benefit school $i$ would have received under the same interventions had the racial/ethnic distribution of school $i$ instead been $\boldsymbol{r}'$. (See \citet{impacts_kusner2019} for the definition of $c_{ir'}$ in the general case.)

Wrapping these two goals (maximizing overall beneficial impact and constraining counterfactual privilege) into an optimization problem, the discriminatory impact problem solves:

\begin{equation}\label{eq:original_schools_objective}
\begin{aligned}
    \underset{z \in \{0, 1\}^m}{\texttt{max}} \quad & \sum_{i=1}^{m} \mathbb{E}[Y^{(i)}(\boldsymbol{r}^{(i)}, \boldsymbol{z}) | R^{(i)} =\boldsymbol{r}^{(i)}, P^{(i)}=p^{(i)}, F^{(i)}=f^{(i)}] \\
    \text{s.t.} \quad & \sum_{i=1}^{m} z^{(i)} \leq b \\
    & c_{ir'} < \tau \quad\quad\quad \forall r' \in R, i \in \{1, \ldots, m\},
\end{aligned}
\end{equation}

\noindent maximizing the expected total impact $Y$, subject to a maximum number of interventions $b$ and constraining counterfactual privilege to remain below $\tau$. We now describe the corresponding disaggregated approach for impact remediation.

\subsection{Disaggregated NYC Schools Example}

In the general case, we would choose our setting for impact remediation based on an existing disparity we have measured. In this case, we have already chosen the setting of the original problem for comparison. Despite the already chosen setting, upon gathering data we see that there is indeed an imbalance in outcome rates $Y$ across groups, with the following observed outcome rates for groups A through G: 

\begin{center}
    \begin{tabular}{c|c|c|c|c|c|c}
        \centering
        A&B&C&D&E&F&G \\ \hline
        0.249&0.243&0.329&0.400&0.355&0.258&0.297
    \end{tabular}
\end{center}

In other words, only about 25\% of students are taking entrance exams for groups A, B, and F, compared to rates as high as 40\% for students in group D.

Observe that, in the original example $c_{ir'}$ requires a counterfactual with respect to racial/ethnic distribution $\boldsymbol{r}$, bringing with it some of the challenges discussed in Section~\ref{sec:social_categories}. Having now gathered disaggregated data directly on this disparity, we are able to take a different approach. To be more precise, we have now measured $Y$ in a disaggregated manner as follows:

\begin{itemize}
    \item $Y^{(i)}_{k}$ = fraction of students in social category $k$ at school $i$ who take college entrance exams (SAT or ACT).
\end{itemize}

\begin{figure}
    \centering
    \fbox{\includegraphics[width=0.36\textwidth]{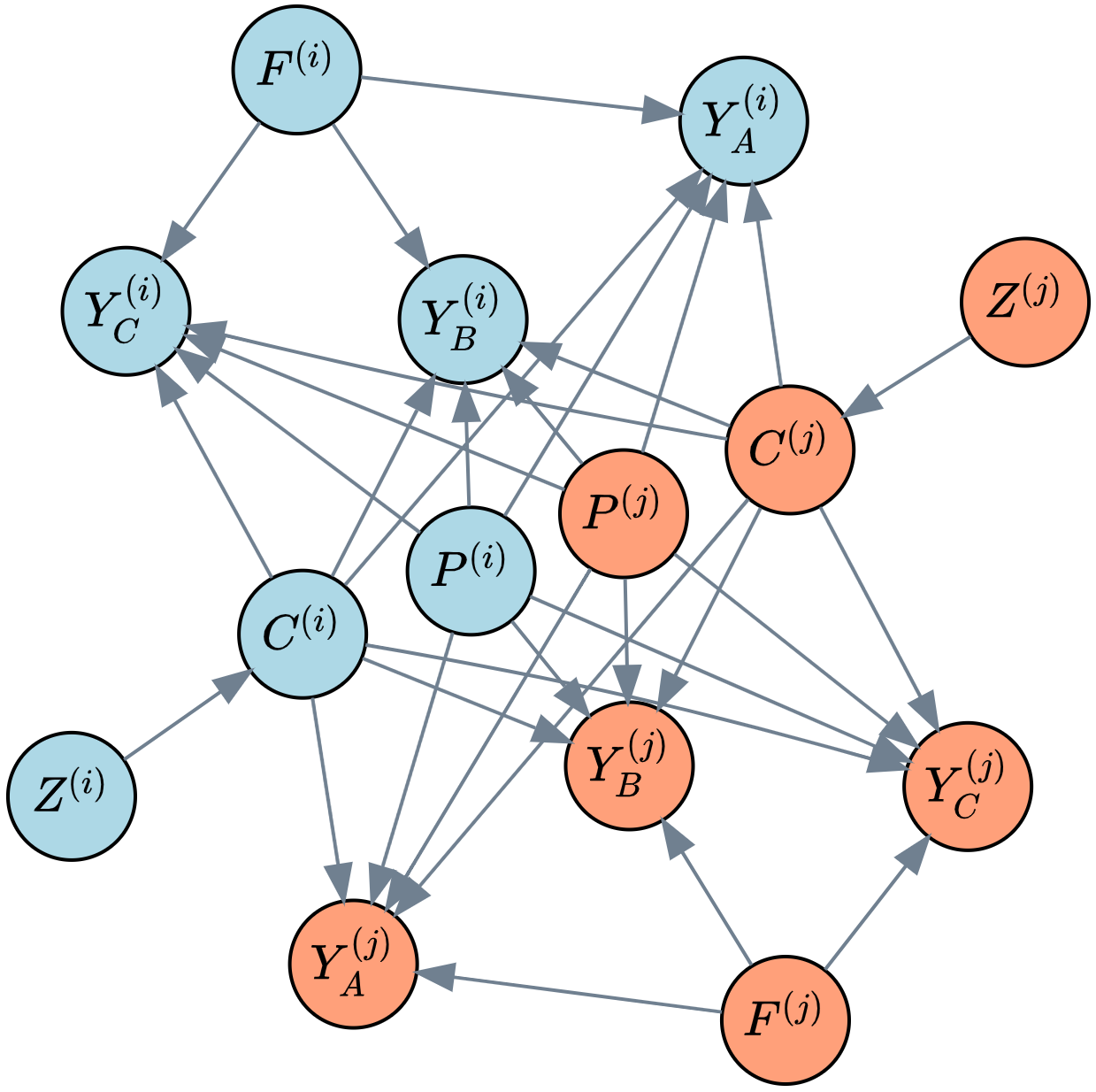}}
    \caption{Disaggregated causal graph structure for the NYC schools example. This image, again, depicts the subset of the graph corresponding to two neighboring schools $i$ and $j$, now with no causal variable for sensitive social attribute $R$ and with the outcome $Y$ disaggregated across social categories A, B, and C.}
    \label{fig:new_schools_graph}
\end{figure}

\begin{figure*}
\centering
\setlength\fboxsep{0pt}
\setlength\fboxrule{0.25pt}
\includegraphics[width=\textwidth]{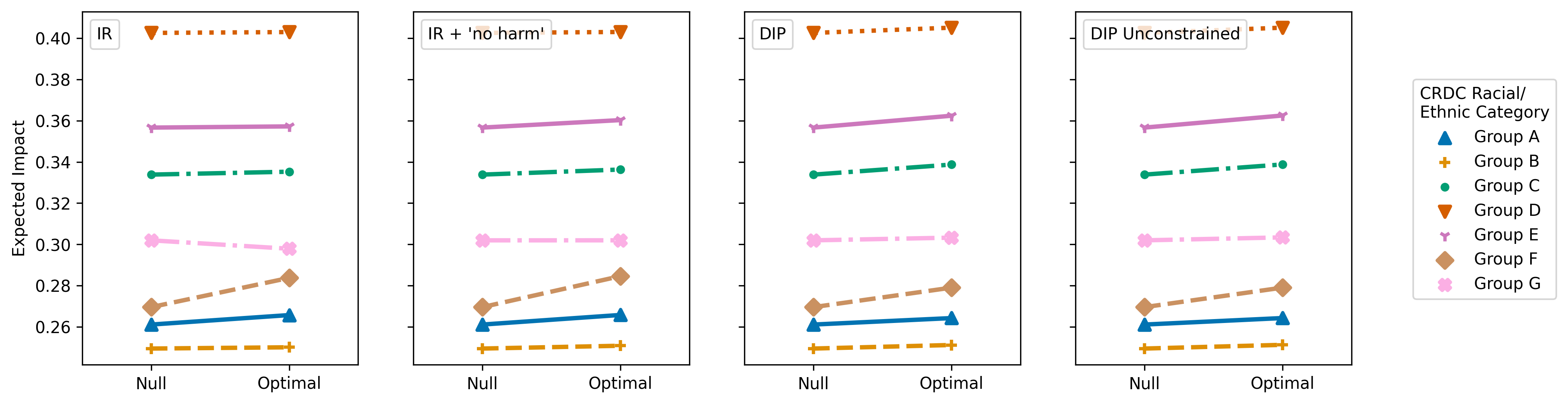}
\caption{Expected impact for each social category before and after performing the interventions recommended by each method --- from left to right: impact remediation (IR), impact remediation with an additional no-harm constraint (IR + `no harm'), the discriminatory impact problem with $\tau=0.567$ (DIP), and the discriminatory impact problem with $\tau$-controlled counterfactual privilege unconstrained (DIP Unconstrained).}
\label{fig:lettered_comparison}
\end{figure*}

Alongside this disaggregation, we no longer model the racial/ethnic distribution $R$ as a causal variable. We maintain analogous assumptions that all observed variables at school $i$ directly influence each impact $Y^{(i)}_{k}$, and that $C^{(i)}, P^{(i)}$ also influence $Y^{(j)}_{k}$ if school $j$ is one of the nearest $K=5$ neighboring schools. Our new disaggregated causal graph structure is shown in Figure \ref{fig:new_schools_graph}. To declutter the diagram, we show the outcome disaggregated over only 3 groups --- A, B, and C. In actuality, there are 7 groups --- A, B, C, D, E, F, and G. In order to make the comparison as direct as possible, we keep roughly the same functional form for our structural equations, this time for each $Y^{(i)}_{k}$ instead of one aggregate $Y^{(i)}$:

\begin{align*}
&\mathbb{E}[Y^{(i)}_{k}(\boldsymbol{z})|R^{(i)} = \boldsymbol{r}^{(i)}, P^{(i)}=p^{(i)}, F^{(i)}=f^{(i)}]\\
&\hspace{0.5cm}= \boldsymbol{\alpha}_k^T\boldsymbol{r}^{(i)} \cdot \underset{j \in N(i)}{\texttt{max}} s(i, j) \left(C^{(j)} \lor C^{(j)}(\boldsymbol{z}) \right)\\
&\hspace{0.5cm}\hspace{0.1cm}+ \boldsymbol{\beta}_k^T\boldsymbol{r}^{(i)} \cdot \underset{j \in N(i)}{\texttt{max}} s(i, j) p^{(j)}\\
&\hspace{0.5cm}+ \boldsymbol{\gamma}_k^T\boldsymbol{r}^{(i)} f^{(i)}\\
&\hspace{0.5cm}+ \boldsymbol{\theta}_k^T\boldsymbol{r}^{(i)}
\end{align*}

now with parameters $\boldsymbol{\alpha}_k, \boldsymbol{\beta}_k, \boldsymbol{\gamma}_k, \boldsymbol{\theta}_k$ for each social category $k$, again, fit via least squares. In our case, we consider the observed presence of Calculus courses under the null intervention, replacing $C^{(j)}(\boldsymbol{z})$ with $C^{(j)} \lor C^{(j)}(\boldsymbol{z})$, because the current presence of courses impacts the current differences in outcomes between groups, which are relevant to our optimization objective. Although we have the same functional form involving $\boldsymbol{r}^{(i)}$, the underlying causal assumption with respect to $R$ has changed --- the relationship is now purely associational, and intervention with respect to $R$ is undefined. Hence, $R$ is no longer part of the causal graph.

We saw several potential objective functions and constraints in Section~\ref{sec:design}. Consider the following choice of disaggregated intervention design for this example, where we measure disparity in expectation as

\begin{align*}
\delta(\boldsymbol{z}) = \sum_{k \neq k'} \left| \mu_k -  \mu_{k'} \right|
\end{align*}

\noindent where

\begin{align*}
\mu_k & = \frac{1}{n_{k}} \sum_{i=1}^{m} n^{(i)}_k \mathbb{E}[Y^{(i)}_{k}(\boldsymbol{z})|R^{(i)} = \boldsymbol{r}^{(i)}, P^{(i)}=p^{(i)}, F^{(i)}=f^{(i)}],\\
\mu_{k'} & =  \frac{1}{n_{k'}} \sum_{i=1}^{m} n^{(i)}_{k'} \mathbb{E}[Y^{(i)}_{k'}(\boldsymbol{z})|R^{(i)} = \boldsymbol{r}^{(i)}, P^{(i)}=p^{(i)}, F^{(i)}=f^{(i)}],
\end{align*}

\noindent and we solve

\begin{equation}\label{eq:new_schools_objective}
\begin{aligned}
    \underset{z \in \{0, 1\}^m}{\texttt{min}} \quad & \delta(\boldsymbol{z}) \\
    \text{s.t.} \quad & \sum_{i=1}^{m} z^{(i)} \leq b
\end{aligned}
\end{equation}

minimizing our measure of expected disparity subject to the same maximum number of interventions $b$. We also solve a second version of this optimization problem with the following additional `no harm' constraint, as described in Section~\ref{sec:design}:

\begin{equation*}
    \begin{split}
        \frac{1}{n_k}  \sum_{i=1}^{m} n^{(i)}_k \left( \mathbb{E}[Y^{(i)}_{k}(\boldsymbol{z})|R^{(i)} = \boldsymbol{r}^{(i)}, P^{(i)}=p^{(i)}, F^{(i)}=f^{(i)}] \right. \\
        \left. - \mathbb{E}[Y^{(i)}_{k}|R^{(i)} = \boldsymbol{r}^{(i)}, P^{(i)}=p^{(i)}, F^{(i)}=f^{(i)}] \right) \geq 0 \ \ \forall k.
    \end{split}
\end{equation*}

\subsection{Results}

In order to solve the optimization problems in both the original and disaggregated approach, we use the same reformulation to a mixed-integer linear-program (MILP) introduced in \cite{impacts_kusner2019}, with only a slight distinction: our alternate objective function requires that we linearize an absolute value, which is automatically handled by the Gurobi optimization package in Python \cite{gurobi}. As the MILP itself is not our focus here, we refer the reader to the specification of the MILP in \citet{impacts_kusner2019} for details on auxiliary variables and an MILP reformulation of the optimization problem in Equation \ref{eq:original_schools_objective} that directly applies to our new optimization problem in Equation \ref{eq:new_schools_objective}.

Solving both optimization problems with our new dataset demonstrates how impact remediation can lead to very different policy recommendations. To capture both ends of the spectrum of solutions for $\tau$-controlled counterfactual privilege, we solve two versions of the original discriminatory impact problem, one with $\tau=0.567$ --- found to be the smallest feasible value of $\tau$ to three decimal places --- and another with infinite $\tau$ (\ie unconstrained $c_{ir'}$). We solve both the discriminatory impact problem (DIP) and impact remediation (IR) with a budget of $b=100$ out of the 490 schools. 

Figure \ref{fig:lettered_comparison} shows, for each method, the expected impact in each group with the optimal intervention compared to the null-intervention. Expected impact under the null-intervention is computed using the observed presence of Calculus courses $C^{(i)}$ while allowing interference. We see gains for many groups in both versions of DIP. In regular IR, we see at least one group's expected impact values decrease rather than increase, a trend remedied by adding the `no harm' constraint. Taking a closer look, Table \ref{tab:impact_comparison_table} shows our measure of expected disparity for each set of interventions, along with the percentage change in expected per-group impact $\frac{1}{n_k} \sum_{i=1}^{m} n^{(i)}_k \mathbb{E}[Y^{(i)}_{k}]$ for each $k \in \{A, B, C, D, E, F, G\}$, and the percentage change in aggregate expected impact across all students $\frac1n \sum_{k=1}^{r} \sum_{i=1}^{m} n^{(i)}_k \mathbb{E}[Y^{(i)}_{k}]$. We see that, with no intervention at all, the expected disparity measure is 1.429. Regular IR is able to lower this value the most, to 1.386, but we observe that this has come at the cost of lower outcome rates for group G. Including the `no harm' constraint avoids this issue, leading to even larger increases for groups that previously saw gains from regular IR, and we still see a decrease in overall disparity. Notice that our disaggregated approach in IR can more directly reveal when the effect of a possible intervention is not universally positive and may cause harm for some sub-population(s), though a full understanding of this possibility would require a full analysis of the causal model fit and coefficients. The last two rows of the table show that, in both instances of DIP, although the aggregate impact is the highest (which is the objective of DIP), the groups that need the least help, with the exception of group B, see their largest gains in impact, while most of the groups with lower outcome rates see smaller gains. The $\tau$-controlled counterfactual privilege constraint has dampened this effect slightly, but has had little effect on disparity --- in both cases, disparity has increased compared to performing no interventions.

\begin{table*}
\centering
\caption{Expected disparity, the percentage change in aggregate expected impact across all students, and the percentage change in expected impact for each social category A-G after performing the interventions recommended by each method.}\label{tab:impact_comparison_table}
\begin{tabular}{llllllllll}
\hline
\multirow{2}{*}{\textbf{Approach}} & \multicolumn{7}{c}{\textbf{\% Change in Impact Per-group}} & \textbf{Aggregate} & \textbf{Disparity} \\ \cline{2-8}
& \textbf{A} & \textbf{B} & \textbf{C} & \textbf{D} & \textbf{E} & \textbf{F} & \textbf{G} & \textbf{\% Impact} & $\boldsymbol{(\delta(z))}$ \\ \hline
No Intervention     & $\pm 0.0$  & $\pm 0.0$ & $\pm 0.0$ & $\pm 0.0$  & $\pm 0.0$ & $\pm 0.0$ & $\pm 0.0$ & $\pm 0.0$ & 1.429 \\
IR                  & +1.76 & +0.24 & +0.42 & +0.10 & +0.16 & +5.26 & --1.35 & +0.657 & \textbf{1.386} \\
IR + `no harm'      & \textbf{+1.78} & +0.54 & +0.74 & +0.11 & +1.02 & \textbf{+5.56} & $\pm 0.0$ & +0.848 & 1.394 \\
DIP, $\tau = 0.567$ & +1.20 & +0.69 & +1.46 & \textbf{+0.63} & +1.61 & +3.50 & +0.43 & +0.953 & 1.435 \\
DIP, Unconstrained  & +1.21 & \textbf{+0.72} & \textbf{+1.48} & \textbf{+0.63} & \textbf{+1.64} & +3.51 & \textbf{+0.47} & \textbf{+0.971} & 1.435 \\
\end{tabular}
\end{table*}

{\bf In summary}, our proposed framework, IR, with the no harm option, is in this case most effective at reducing disparities where it matters most --- prioritizing groups with lower measured outcomes (such as A and F) and decreasing disparity overall.

\begin{figure}
    \centering
    \includegraphics[width=0.5\textwidth]{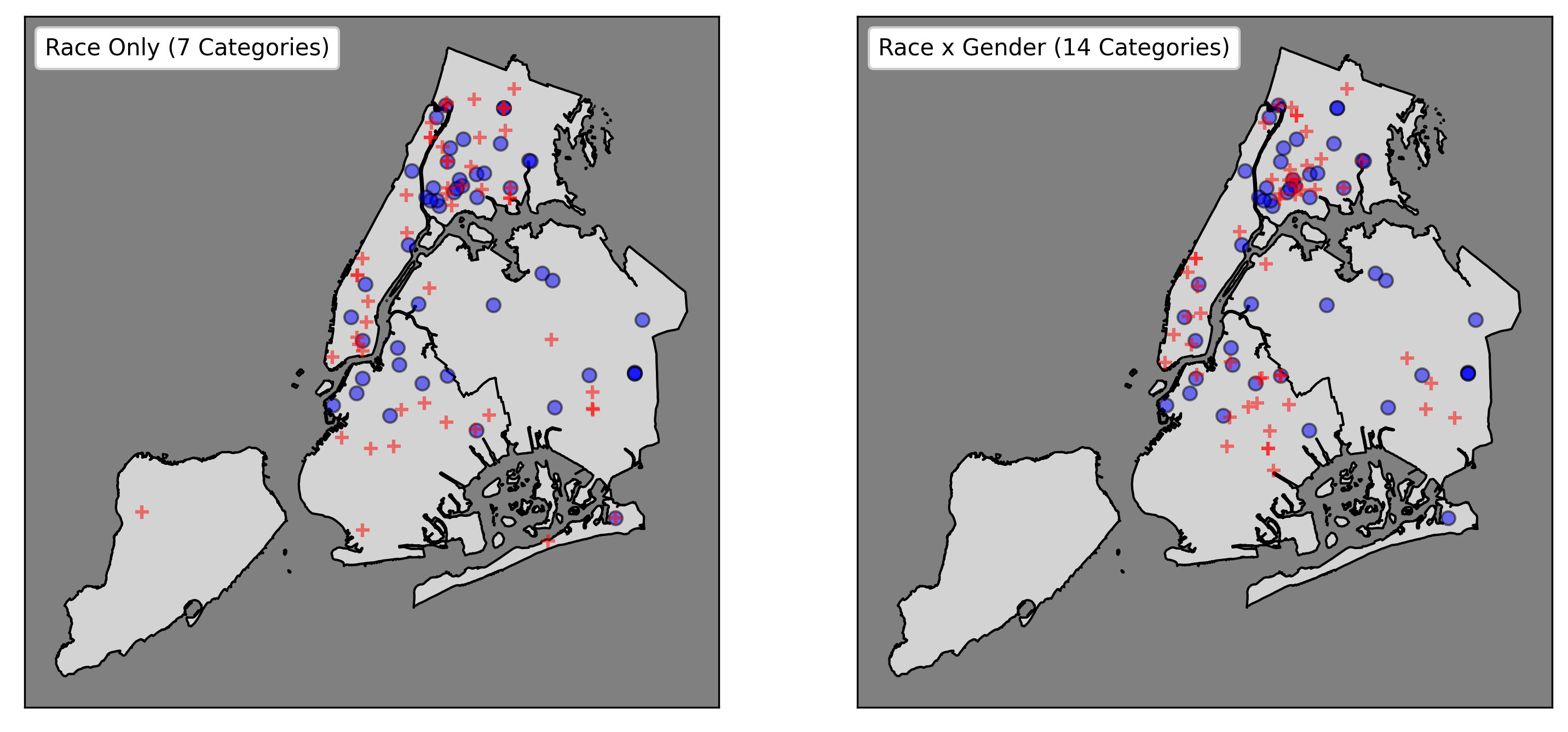}
    \caption{Comparison of which 100 schools are chosen by impact remediation using two different social partitions of the same population. On the left, we partition the population based on seven CRDC racial/ethnic categories. On the right, we partition the population based on the Cartesian product of seven CRDC racial/ethnic categories and two gender categories, or 14 groups overall. Schools selected in both problems are indicated by blue dots. Those unique to each problem are indicated by red plus symbols.}
    \label{fig:map_comparison}
\end{figure}

\begin{figure}
    \centering
    \includegraphics[width=0.5\textwidth]{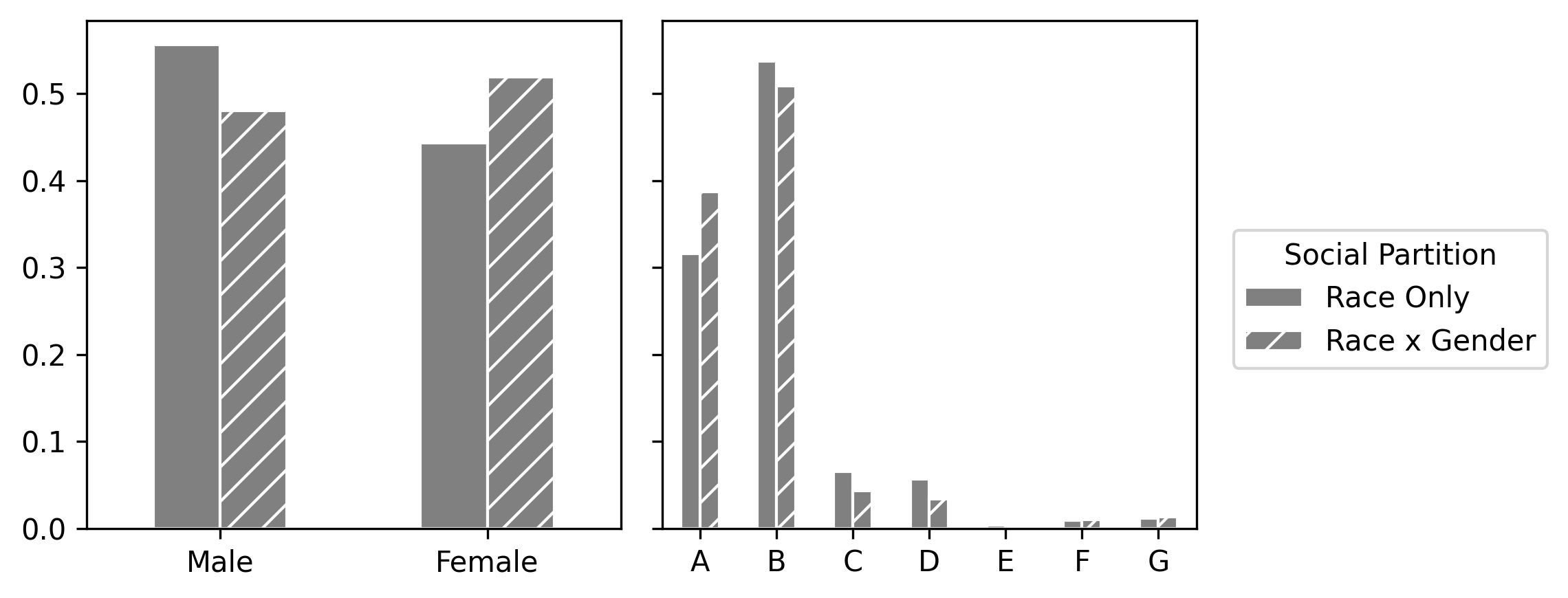}
    \caption{Comparison of the different gender and racial distributions across schools chosen by two impact remediation settings, each using a different social partition of the same population. Students tabulated for each social partition are those unique to each impact remediation setting \ie students corresponding to the red plus symbols in Figure \ref{fig:map_comparison}.}
    \label{fig:hist_comparison}
\end{figure}

Although our example here is stylized compared to a full real-world application problem, it demonstrates how the impact remediation framework's focus on disparity rather than on utility as an objective can lead it to very different policy recommendations.
Observe also that our decision problem depends directly on what partition of the population we consider disparity across. Figure \ref{fig:map_comparison} shows visually how different an allocation of optimal schools can be for impact remediation if we consider racial/ethnic categories only, versus considering the intersection of both racial/ethnic categories and gender categories. Considering the intersection of these categories has significantly changed which schools are chosen for additional funding. Figure \ref{fig:hist_comparison} shows the corresponding distributions across gender and racial/ethnic categories for the schools unique to each version of impact remediation. Including gender in our social partition (indicated with stripes) has altered both of these distributions --- across gender and across racial/ethnic groups. For example, in the racial/ethnic distribution, the fractions of group A have shifted noticeably. Impact remediation including gender also targets a larger fraction of students in the Female group. 

{\bf In summary,} our results demonstrate how the policy recommendations of impact remediation will depend not only on our chosen measure of disparity or optimization constraints, but also on which groups we consider.

\section{Conclusion and Open Problems}
\label{sec:conclusion}

In this work, we have formalized the impact remediation problem, which leverages causal inference and optimization to focus on modeling measured disparities and mitigating them through real-world interventions. Our framework flexibly allows us to consider the role of social categories, while relaxing the typical modeling assumptions that their use in causal modeling requires. We demonstrated impact remediation with a real-world case study, and saw how it can decrease disparity and how it differs from existing approaches both qualitatively and quantitatively.

The framework is general enough to accommodate a broad range of problems, and can be used to explore additional questions beyond those we have covered here. For example, we know that each specific case of impact remediation will be different. Having a framework around these problems could allow us to explore patterns across impact remediation in different settings and/or with different causal modeling techniques. In the special case without interference, we can connect to the growing literature on causal inference for heterogeneous treatment effects and optimal policy recently reviewed by \citet{athey2019machine}. Understanding potentially heterogeneous treatment effects can help us better understand why particular units are chosen for intervention, and help policymakers decide which policy to adopt.

Another area where the impact remediation framework affords future exploration is in modeling intersectionality, and considering in-group fairness with multi-level sensitive attributes that potentially influence each other in unknown ways \cite{mhasawade_causal_2020, Yang2020CausalIF, Foulds2020AnID, balanced_ranking_yang_2019}.
Consider, for example, racial categories and socioeconomic indicators such as income. For the reasons discussed in Section~\ref{sec:social_categories}, these two concepts could simultaneously be incorporated into impact remediation to make more nuanced policy decisions without requiring specific causal assumptions on the nature of the relationship between race and socioeconomic status.

\begin{acks}
This research was supported in part by \grantsponsor{NSF}{National Science Foundation (NSF)}{https://www.nsf.gov} awards No. \grantnum{NSF}{1922658}, \grantnum{NSF}{1934464}, \grantnum{NSF}{1926250}, and \grantnum{NSF}{1916505}.
\end{acks}

\bibliographystyle{ACM-Reference-Format}
\bibliography{references}


\end{document}